\documentclass[letterpaper, 10 pt, conference]{ieeeconf}  

\IEEEoverridecommandlockouts                              

\usepackage{graphics} 
\usepackage{epsfig} 
\usepackage{mathptmx} 
\usepackage{times} 
\usepackage{amsmath} 
\usepackage{amssymb}  
\usepackage{graphicx}

\usepackage{algorithm}
\usepackage{algpseudocode}

\usepackage{subcaption}

\usepackage[hidelinks]{hyperref} 

\usepackage{booktabs} 

\usepackage{pgfplots}
\pgfplotsset{compat=1.15,
	legend style={font=\footnotesize},
}
\usepackage{tikz}
\usepackage{tikzscale}
\usepgfplotslibrary{fillbetween}

\usepgfplotslibrary{patchplots}

\usepackage{stfloats} 

\usepackage{bm}
\usepackage{optidef}

\title{\LARGE \bf
Convex Maneuver Planning for Spacecraft Collision Avoidance
}

\author{Fausto Vega$^{1}$, Jon Arrizabalaga$^{1}$, Ryan Watson$^{2}$, and Zachary Manchester$^{1}$
\thanks{$^{1}$Fausto Vega, Jon Arrizabalaga, ands Zachary Manchester are with the Robotics Institute at Carnegie Mellon University,
        Pittsburgh, PA 15213, USA
        {\tt\small \{fvega, jarrizab, zmanches\}@andrew.cmu.edu}}%
\thanks{$^{2}$Ryan Watson is with Albedo Space,
       Broomfield, CO 80020, USA
       {\tt\small rwatson@albedo.com}}%
}

\begin{document}

\maketitle 
\thispagestyle{empty}
\pagestyle{empty}

\begin{abstract}

Conjunction analysis and maneuver planning for spacecraft collision avoidance remains a manual and time-consuming process, typically involving repeated forward simulations of hand-designed maneuvers. With the growing density of satellites in low-Earth orbit (LEO), autonomy is becoming essential for efficiently evaluating and mitigating collisions. 
In this work, we present an algorithm to design low-thrust collision-avoidance maneuvers for short-term conjunction events. We first formulate the problem as a nonconvex quadratically-constrained quadratic program (QCQP), which we then relax into a convex semidefinite program (SDP) using Shor's relaxation. We demonstrate empirically that the relaxation is tight,  which enables the recovery of globally optimal solutions to the original nonconvex problem. Our formulation produces a minimum-energy solution while ensuring a desired probability of collision at the time of closest approach. Finally, if the desired probability of collision cannot be satisfied, we relax this constraint into a penalty, yielding a minimum-risk solution. We validate our algorithm with a high-fidelity simulation of a satellite conjunction in low-Earth orbit with a simulated conjunction data message (CDM), demonstrating its effectiveness 
in reducing collision risk. 

\end{abstract}

\section{INTRODUCTION}


The growing density of low-Earth orbit (LEO) traffic has made collision avoidance a pressing challenge for satellite operators. Rapid advances, including reusable launch vehicles, spacecraft miniaturization \cite{woellert2011cubesats}, and large constellations like Starlink, have accelerated the pace of satellite deployment. As a result, LEO has become increasingly congested, raising the likelihood of conjunction events and imposing greater operational demands on satellite operators. To address these challenges, autonomous maneuver planning is needed to enable timely and scalable collision avoidance for safer and more efficient space-traffic management. 

Despite the growing need for autonomy, conjunction analysis and maneuver planning remain largely manual and time consuming. Current operations typically involve repeated forward simulations of hand-designed maneuvers, adjusting them iteratively until safety criteria are met. Tools exist to assess the effect of these candidate maneuvers \cite{alfano2005collision}, but the process remains labor intensive and difficult to scale. 
Modeling maneuver planning as an optimal control problem provides a structured framework for balancing safety, fuel efficiency, and operational constraints. However, the problem is inherently nonconvex due to nonlinear orbital dynamics and collision-avoidance constraints. One option is to solve the problem directly as nonlinear program \cite{kelly2017introduction}, but this local method depends heavily on the quality of the initial guess. Sequential convex programming (SCP) is an example of this approach that approximates the problem through a sequence of convex subproblems \cite{mao2017successive}. SCP can converge quickly when initialized near the optimal solution, but it does not guarantee feasibility with respect to the original nonconvex problem and remains sensitive to initialization and hyperparameter tuning \cite{malyuta2022convex}. Convex relaxations provide another approach, reformulating the problem so that the nonconvex constraints become convex \cite{accikmecse2010lossless, yang2022certifiably}. Under certain conditions, this method can yield the global optimum, and this is the approach we adopt.

In spacecraft collision avoidance, the linear relationship between an impulsive maneuver and a displacement at closest approach can address the nonconvex collision constraint \cite{bombardelli2014collision}. However, this method does not optimize for fuel --- a critical resource that constrains mission lifetime. An alternative is an iterative projection–linearization algorithm, which approximates the nonconvex probability-of-collision (PoC) constraint as a half-space and solves a sequence of convex subproblems \cite{armellin2021collision}. While effective in generating impulsive maneuvers, this method depends on a good initial guess and lacks global optimality guarantees.
 
Other approaches include analytical methods, such as a closed‑form solution for the burn time required to achieve a desired PoC at the time of closest approach (TCA) \cite{strobel2024analytical}. This formulation provides a direct computation for burn timing but does not optimize maneuver direction and assumes along‑track burns for the entire maneuver. To avoid fuel expenditure, collision‑avoidance maneuvers using atmospheric drag have also been explored \cite{gaglio2025optimal}. However, the trade off is a loss in orbital altitude and shorter mission lifetime. 

While prior methods represent important progress toward autonomous maneuver planning, none simultaneously achieves global optimality, handles general low‑thrust maneuvers, and remains applicable to short‑term conjunction events.
To address this combination of requirements, we propose a globally optimal algorithm for designing low‑thrust collision‑avoidance maneuvers applicable to short‑term conjunction events. Our specific contributions include:
\begin{enumerate}
\item A convex reformulation of the nonconvex PoC constraint and thrust-lower-bound constraint using Shor’s relaxation.
\item A minimum-energy trajectory-optimization framework that generates low‑thrust maneuvers to achieve a target PoC at the TCA.
\item High‑fidelity LEO conjunction simulation, demonstrating the tightness of the proposed formulation and the effectiveness of the overall approach.
\end{enumerate}

The remainder of this paper is organized as follows: Section \ref{background} introduces convex relaxation concepts and provides background on spacecraft conjunction analysis. Section \ref{Method} derives our convex formulation. Section \ref{experiments-results} presents experiments and results for a simulated LEO conjunction. Finally, Section \ref{conclusions} summarizes our conclusions and discusses future work.



\section{BACKGROUND}
\label{background}

This section provides a brief review of convex relaxations and spacecraft conjunction analysis. We refer the reader to \cite{luo2010semidefinite} and \cite{krage2023nasa} for more comprehensive treatments.

\subsection{Convex Relaxations}
\label{cvx-relaxations}
For an optimization problem to be convex, equality constraints must be affine, and both the objective function and inequality constraints must be convex \cite{boyd2004convex}. Convex relaxations reformulate nonconvex problems into convex ones that can be solved efficiently to global optimality. When certain conditions are satisfied, the solution to the convex problem is also the globally optimal solution to the original nonconvex problem. This approach has been successfully applied across various domains including robotics \cite{marcucci2023motion} \cite{carlone2018convex} and aerospace applications \cite{acikmese2007convex}, most notably the rocket soft-landing problem \cite{acikmese2013lossless}. 

The effectiveness of a convex relaxation depends strongly on the problem's structure. In this work, our problem of interest is a quadratically-constrained quadratic program (QCQP) --- an optimization problem where both the cost function and constraints are quadratic. In general, QCQP's are nonconvex and NP-hard, making them challenging to solve to global optimality \cite{d2003relaxations}. Convex relaxation techniques can be used to compute a tractable lower bound on the optimal value $f^*$. 

To understand this idea better, consider the following nonconvex QCQP:
\begin{equation} 
\begin{aligned}
f^\star =\; & \min_{x \in \mathbb{R}^n} \quad x^\top Q x + d^\top x\\
\text{s.t.}\; & x^\top A x = b ,
\label{qcqp}
\end{aligned}
\end{equation}
where the matrices $Q$ and $A$ are not necessarily positive definite. Shor's relaxation \cite{shor1987quadratic} reformulates \eqref{qcqp} as a convex semidefinite program (SDP), which can be solved in polynomial time using interior-point methods \cite{vandenberghe1996semidefinite}. The key idea is to define the \emph{moment matrix} $M = zz^\top$, where $z = [1 \quad x]$:
\begin{equation}
M = \begin{bmatrix}
1 & x^\top \\ 
x & xx^\top
\end{bmatrix}
\end{equation}
 We also define a set of (linear) functions $L$ to pick out the $x$ and $x x^\top$ terms from $M$ ($x = L_x(M), \ xx^\top = L_{xx}(M)$). These functions consist of a left and right matrix multiplication on $M$ to select the desired terms. Rewritten in terms of $M$ by applying the trace property $\operatorname{tr}(x^\top Q x) = \operatorname{tr}(Q xx^\top)$, the objective and constraint functions in \eqref{qcqp} become linear, but we now have the non-convex equality constraint $M=zz^\top$ which is equivalent to $M \succeq 0$ and $\operatorname{rank}(M) = 1$:
\begin{equation} 
\begin{aligned}
f^\star =\; & \min_{M} \quad \operatorname{tr}(QL_{xx}(M)) + d^\top L_x (M) \\
\text{s.t.}\; & \operatorname{tr}(AL_{xx}(M)) = b \\
              &M \succeq 0, \quad \operatorname{rank}(M) =1  \\
\label{sdp-rank}
\end{aligned}
\end{equation}
We can relax \eqref{sdp-rank} by dropping the nonconvex rank constraint, which corresponds to relaxing $M -zz^\top = 0$ into $M - zz \succeq 0$.
This allows us to arrive at the following SDP:
\begin{equation} \label{eq:sdp-opt}
\begin{aligned}
p^\star =\; & \min_{M} \quad \operatorname{tr}(QL_{xx}(M)) + d^\top L_x (M) \\
\text{s.t.}\; & \operatorname{tr}(AL_{xx}(M)) = b, \quad M \succeq 0 ,\\
\end{aligned}
\end{equation}  
which is convex and provides a strict lower bound on the original problem in \eqref{qcqp}. 
We say that the relaxation \eqref{eq:sdp-opt} is \emph{tight} if its solution matches the solution to the original problem \eqref{qcqp}. In this case, tightness corresponds to the optimal moment matrix $M^*$ having rank one (i.e. it should correspond to the outer product $zz^\top$). This rank condition can be checked directly after solving the SDP, providing a certificate of global optimality. 
  
If $M^*$ is not rank one, the relaxation is not tight, and $p^*$ lower bounds $f^*$ ($p^* \leq f^*$).  However, in some cases, a rank-one approximation can  be extracted from $M^*$ using  techniques such as an eigenvalue decomposition \cite{luo2010semidefinite}. This approximation can then serve as an initial guess for a nonlinear programming (NLP) solver directly applied to \eqref{qcqp} \cite{teng2023convex}. 

In summary, convex relaxations provide a way to obtain certified globally optimal solutions (if they are tight) or high-quality approximate solutions for nonconvex problems. We apply these techniques to spacecraft collision avoidance, where the problem formulation and constraints arise from spacecraft conjunction analysis.


\subsection{Spacecraft Conjunction Analysis}

In spacecraft conjunction analysis, the probability of collision ($P_c$) is a critical metric used to ensure safe operations in congested orbital environments. Satellite operators submit a formatted ephemeris with an associated covariance matrix to designated screening organizations. These submissions are analyzed to assess potential conjunctions with other space objects. If the computed $P_c$ exceeds a predefined threshold, a Conjunction Data Message (CDM) is generated, issuing a proximity warning to the operator and prompting potential mitigation actions, such as collision avoidance maneuvers \cite{krage2023nasa}.

A CDM includes the estimated state vectors and associated covariance matrices of both the primary and secondary satellites at the TCA. In this work, we assume that the secondary satellite remains passive, while the primary satellite executes an avoidance maneuver to reduce the probability of collision at the TCA.




\subsubsection{Probability of Collision}
\label{PoC}
\begin{figure*}[t]
  \centering
    \includegraphics[width=1.75\columnwidth,trim=4.5cm 5.8cm 1.5cm 2.5cm, 
    clip]{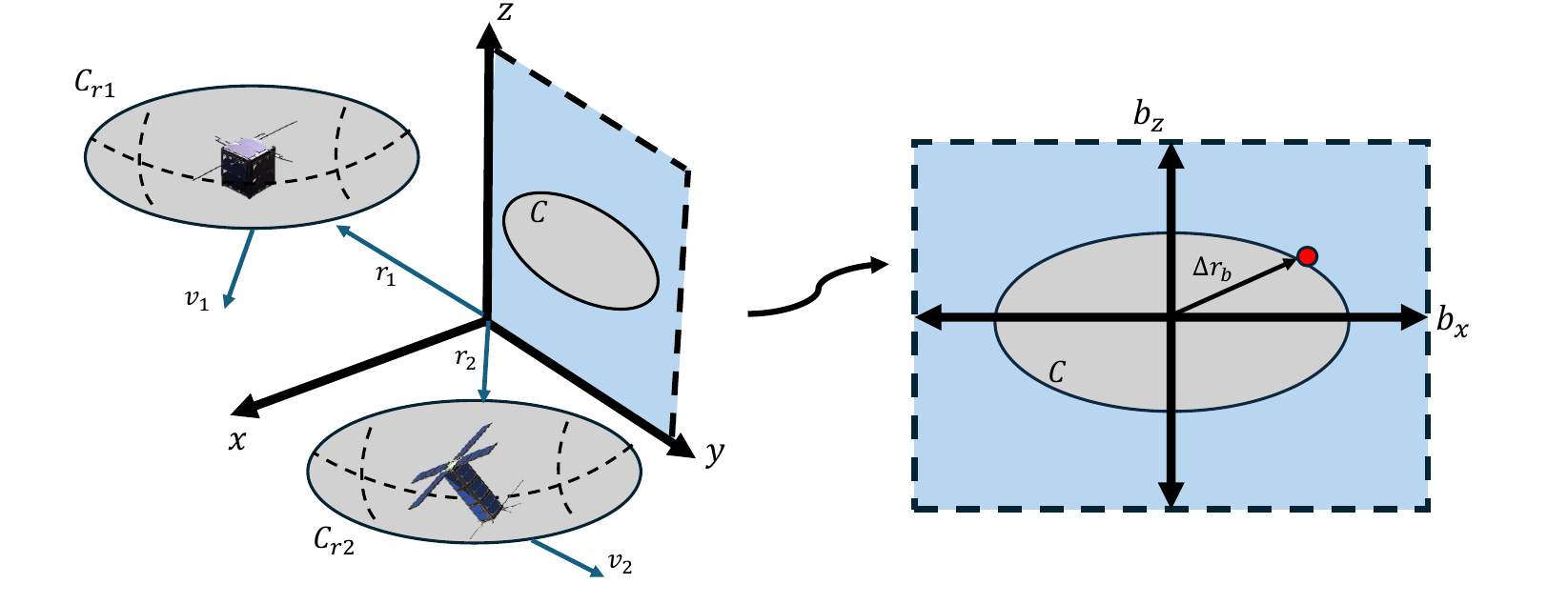}
    \caption{The probability-of-collision constraint accounts for the position uncertainties of both spacecraft, $C_{r1}$ and $C_{r2}$, combined and projected onto the b-plane --- the plane perpendicular to the relative velocity vector. The b-plane is then centered on the secondary spacecraft, yielding an ellipse centered on the secondary spacecraft. Points inside this ellipse have high $P_c$, so the feasible region is the blue-shaded, nonconvex set outside the ellipse. The red point marks the relative position $\Delta r_b$ of the primary spacecraft with respect to the secondary spacecraft in the b-plane.}
    \label{ellipsoid-constraint}
\end{figure*}
The probability of collision is the likelihood that the relative position vector between two spacecraft in the Earth-centered inertial (ECI) frame $\Delta r$ lies within a collision sphere of radius $R_{HBR}$ (sum of their hard-body radii). It is defined as the integral of a 3D Gaussian probability density function (PDF) of $\Delta r$ over this sphere, but this integral lacks a closed-form solution and is unsuitable for direct trajectory optimization. 

We adopt the short-term encounter assumption \cite{pavanello2025cammary}, where high relative velocity at the TCA ($\Delta v$) implies:
\begin{equation}
\Delta r^\top \Delta v = 0
\label{dot-product-condition}
\end{equation}
This orthogonality condition means that, at the TCA, the relative position lies entirely in the encounter plane (b-plane), the plane perpendicular to the relative velocity vector. This allows the 3D collision integral to be reduced to 2D \cite{foster1992parametric}. The b-plane basis is defined as: 
\begin{equation}
    b_y = \frac{\Delta v}{||\Delta v||_2},\ 
    b_z = \frac{\Delta r \times \Delta v}{||\Delta r \times \Delta v||_2},\
    b_x = b_y \times b_z \\
\end{equation}
with the transformation matrix $\tilde{R}$ to transform from the ECI frame to the 2D b-plane: 


\begin{equation}
    R_{ECI}^{ENC} = \begin{bmatrix}
    b_x^\top & b_y^\top & b_z^\top 
    \end{bmatrix},\ 
    E = 
\begin{bmatrix}
1 & 0 & 0 \\
0 & 0 & 1 
\end{bmatrix},\ 
    \tilde{R} = E R_{ECI}^{ENC} \\
\end{equation}

Assuming constant b-plane geometry and a uniform 2D Gaussian PDF over the collision circle, the PoC constraint can be expressed as \cite{armellin2021collision,alfriend1999probability}: 
\begin{equation}
\begin{aligned}
    \Delta r_b^\top C^{-1} &\Delta r_b \geq p \\
    p = \ln(\frac{R_{HBR}^4}{4 P_c^2 \det(C)})&, \ C = \tilde{R}(C_{r1} + C_{r2})\tilde{R}^\top\\
    \label{pc-constraint-nonconvex}
\end{aligned}
\end{equation}
where ($\Delta r_b = \tilde R \Delta r$) is the relative position projected onto the b-plane, $C$ is the sum of the position covariance ($C_{q1}, C_{q2}$) of both spacecraft projected onto the b-plane using $\tilde{R}$, 
and $P_c$ is the desired probability of collision. The equality in the nominal PoC formulation from \cite{armellin2021collision} is relaxed to an inequality, since achieving the desired probability threshold is sufficient for feasibility, but a lower collision probability is always preferable. This relaxation is favorable for solver convergence. The feasible region is inherently nonconvex. Geometrically, this constraint is depicted in Fig. \ref{ellipsoid-constraint}. 

\section{Method}
\label{Method}
We now present
our trajectory-optimization formulation and detail the relaxation process for reformulating it into a convex problem. 

\subsection{System Dynamics}
\label{dynamics}
The primary spacecraft state $x \in {\mathbb{R}}^6$ consists of a position $r \in {\mathbb{R}}^3$ and a velocity $v \in {\mathbb{R}}^3$ in the ECI frame and depends on time $t \in \mathbb{R}^+$. We model the continuous dynamics $f(x, t)$ using SatelliteDynamics.jl \cite{duncan_eddy_satellitedynamicsjl_nodate}, an open source orbital dynamics package. We include a tenth-order gravity model as well as perturbations from atmospheric drag, solar radiation pressure, and third-body perturbations from the Sun and Moon.  We model the control input $u \in {\mathbb{R}}^3$ as an additive acceleration in the continuous dynamics $\dot{x} = f(x, t) + u$. We then obtain a nonlinear discrete-time dynamics model $x_{k+1} = f_d(x_k, u_k, t_k)$ by integrating the continuous dynamics using a Runge-Kutta 7/8 integrator from DifferentialEquations.jl \cite{rackauckas2017differentialequations}.

\subsection{Optimization-Based Collision Avoidance}


We formulate the collision-avoidance maneuver planner as a discrete nonlinear trajectory-optimization problem that minimizes the squared L2 norm of the control inputs $u_k$:

\begin{subequations}\label{eq:nlp}
	\begin{alignat}{3}
    \min_{x_{1:N},\,u_{1:N-1}} J &= \sum_{i=1}^{N-1} ||u_k||^2_2\label{eq:quadratic_nlp_cost}\\
	 \text{s.t.}\quad &x_{k+1} = f_d(x_k, u_k), \ x_1 = x_i, \label{eq:nlp_eq_constr}\\\
    &||u_k||_2 \leq u_b,\ \eqref{pc-constraint-nonconvex} \label{eq:nlp_ineq_cons}
	\end{alignat}
\end{subequations}
subject to a set of constraints. This ``minimum energy'' control approach \cite{parrish2017low} encourages smooth, continuous thrust profiles, making it well-suited for low-thrust propulsion systems. The equality constraints in Eq. \eqref{eq:nlp_eq_constr} include both the discrete dynamics function described in Section \ref{dynamics} and an initial state ($x_i$) constraint. The inequality constraints in Eq. \eqref{eq:nlp_ineq_cons} include the PoC constraint described in Section \ref{PoC} and an upper bound on the control inputs. The parameter $N$ represents the number of discrete time steps along the planned trajectory and we assume the TCA occurs at timestep $N$.

This problem is nonconvex due to the nonlinear dynamics equality constraint and the nonconvex collision-avoidance inequality constraint. As a result, its solution depends on the quality of the initial guess given to a local solver, which is not guaranteed to converge and may only yield a locally optimal solution. In the following sections, we describe how we convexify the problem to enable the computation of a globally optimal solution without an initial guess.

\subsection{Convexifying the Collision-Avoidance Problem}

We first tackle the nonconvexity from the nonlinear dynamics. 

\subsubsection{Linearized Dynamics}
To convexify Eq. \eqref{eq:nlp_eq_constr}, we begin by discretizing the reference uncontrolled orbital trajectory of the primary satellite into $N$ knot points and linearizing the nonlinear discrete-time dynamics from Section \ref{dynamics} about each reference point $\bar{x}_k$ using a first-order Taylor expansion. This provides a local approximation of the true dynamics and results in a set of linearized error-state dynamics:
\begin{equation}
\begin{aligned}
\Delta x_{k+1} \approx &A_k\Delta x_k + B_k  u_k , \\
A_k =  \frac{\partial f_d}{\partial x_k}\bigg|_{(\Bar{x}_k, u_k)}, \ 
B_k =& \frac{\partial f_d}{\partial u_k}\bigg|_{(\Bar{x}_k, u_k)}, \ \Delta x_k = x_k - \Bar{x}_k, \\
\end{aligned}
\label{linear-dynamics}
\end{equation}
where $A_k$ and $B_k$ are the discrete dynamics Jacobians evaluated along the reference trajectory at all timesteps $k$.

\subsubsection{PoC Constraint}
Next, to make the PoC constraint convex, we transform the nonlinear program in \eqref{eq:nlp} into a convex semidefinite program (SDP). We apply Shor's relaxation, described in Section \ref{cvx-relaxations} as follows: First, we define vectors $z_k$ at each timestep $k$ consisting of the delta state ($\Delta x_k$) and control input ($u_k$):

\begin{equation}
z_k = \begin{bmatrix}
    1 \\
    \Delta x_k \\ 
    u_k
\end{bmatrix}_{\forall \ k=\{1:N-1\}}, \ \
z_N = \begin{bmatrix}
    1 \\
    \Delta x_N \\ 
\end{bmatrix}
\end{equation}
We then define the moment matrices ($M_k$):
\begin{equation}
\begin{aligned}
    M_k &= z_kz_k^\top = \begin{bmatrix}
    1 & \Delta x_k^\top & u_k^\top \\ 
    \Delta x_k & \Delta x_k \Delta x_k^\top & \Delta x_ku_k^\top \\
    u_k & u_k\Delta x_k^\top & u_ku_k^\top
    \end{bmatrix}_{\forall \ k=\{1:N-1\}} \\
    M_N &= z_Nz_N^\top = \begin{bmatrix}
    1 & \Delta x_N^\top \\ 
    \Delta x_N & \Delta x_N \Delta x_N^\top \\
    \end{bmatrix} \\
\end{aligned}
\end{equation}
Note that all the $M_k$ matrices are positive semidefinite:
\begin{equation}
\begin{aligned}
    \{M_k &\succeq 0\}_{\forall \ k=\{1:N\}} \\
    \label{psd-M}
\end{aligned}
\end{equation}
Next, we define a set of (linear) functions $L$ to pick out certain terms from $M_k$. This allows us to rewrite the nonconvex problem in \eqref{eq:nlp} as a convex SDP in terms of $M$:
\begin{equation}
\begin{aligned}
 &\Delta x_k = L_x (M_k), \quad \Delta x_k \Delta x_k^\top = L_{xx} (M_k), \ u_k = L_u (M_k)\\
 &u_k u_k^\top = L_{uu} (M_k), \quad \Delta x_k u_k^\top = L_{xu} (M_k), \ \forall \ k=\{1:N-1\} 
\end{aligned}
\label{coeff_matrices}
\end{equation}

\begin{equation}
\Delta x_N = L_{xN} (M_N) \quad  \Delta x_N \Delta x_N^\top = L_{xxN} (M_N) \\
\label{coeff_matrices_N}
\end{equation}
We can express the nonconvex PoC constraint from \eqref{pc-constraint-nonconvex} as a linear function of $M_N$. This constraint encodes a desired probability of collision at the terminal state since we assume the TCA occurs at timestep $N$. Since $r_b = \tilde{R} (\bar{x}_N + \Delta x_N - x_s)$, where $x_s$ is the position of the secondary spacecraft at the TCA, we can express \eqref{pc-constraint-nonconvex} as the following: 
\begin{equation}
(\tilde{R} (\bar{x}_{N} + \Delta x_N - x_{s}))^\top C^{-1}(\tilde{R}(\bar{x}_{N} + \Delta x_N - x_{s})) \geq p
\label{terminal-pc-constraint}
\end{equation} 

By expanding \eqref{terminal-pc-constraint} and using properties of the trace operator, we obtain the following simplified nonconvex constraint. 
\begin{equation}
\operatorname{tr}(C^{-1}  b) + \operatorname{tr}(C^{-1} c(\Delta x_N)) \geq p
\label{nonconvex-pc-constraint-v2}
\end{equation}
where, 
\begin{equation}
\begin{aligned}
b &= \tilde{R}(\bar{x}_N \bar{x}_N^\top - \bar{x}_N x_s^\top  -x_s \bar{x}_N^\top + x_s x_s^\top)\tilde{R}^\top \\
c(\Delta x_N) &= \tilde{R}(\bar{x}_N \Delta x_N^\top  + \Delta x_N \bar{x}_N^\top+ \Delta{x}_N \Delta{x}_N^\top  - \Delta x_N x_s^\top - x_s \Delta x_N^\top)\tilde{R}^\top \
\end{aligned}
\label{nonconvex-pc-constraint-v2-constants}
\end{equation}


Using the functions from \eqref{coeff_matrices_N}, we can obtain any of the elements of $M_N$ and express the constraint in \eqref{nonconvex-pc-constraint-v2} as a function of $M_N$. Using these matrices, $c$ becomes linear in variable $M_N$, which allows the nonconvex PoC constraint in \eqref{nonconvex-pc-constraint-v2} to be expressed as the following linear inequality constraint: 

\begin{equation}
    \operatorname{tr}(C^{-1}b) + \operatorname{tr}(C^{-1}c(M_N)) \geq p ,
    \label{sdp-pc-constraint}
\end{equation}
where, 
\begin{multline}
c(M_N) = \tilde{R}\big(\bar{x}_N(L_{xN}(M_N))^\top + L_{xN} (M_N) \bar{x}_N^\top \\
\quad + L_{xxN} (M_N) - L_{xN} (M_N) x_s^\top - x_s (L_{xN}(M_{N}))^\top\big)\tilde{R}^\top
\end{multline}

The variable $M_K$ contains the outer product term $\Delta x_k \Delta x_k^\top$ as part of its block structure. To improve the tightness of the convex relaxtion, we introduce an additional redundant constraint that enforces the consistency between the outer product terms and the linear dynamics. This constraint is given by: 

\begin{equation}
\begin{aligned}
L_{xx} (M_{k+1}) &= A_k L_{xx}(M_k)A_k^\top + A_k L_{xu} (M_k) B_k^\top \\
&+ (A_k L_{xu} (M_k) B_k^\top)^\top + B_k L_{uu}(M_k)B_k^\top \\
\end{aligned}
\label{quadratic-dynamics-sdp}
\end{equation}

With this formulation, quadratic equality constraints and certain nonconvex inequality constraints, can be expressed as linear functions of $M$. For instance, a nonconvex lower bound $b_l$ on the L2 norm of the control inputs can be reformulated as:

\begin{equation}
\operatorname{tr}(L_{uu}(M_k)) \geq b_l^2
\label{lower-bound}
\end{equation}
In the next section, \eqref{lower-bound} is incorporated into the second example to demonstrate the SDP method’s advantage over directly solving the original nonlinear program.

We have now transformed the original nonlinear, nonconvex collision‑avoidance problem in \eqref{eq:nlp} into a formulation that can be solved efficiently with global optimality guarantees. The resulting convex semidefinite program, presented in \eqref{eq:sdp}, defines the core component of our collision-avoidance maneuver-planning method:
\begin{subequations}\label{eq:sdp}
    \begin{alignat}{5}
    \min_{M_{1:N}} \quad & J = \sum_{i=1}^{N-1} \operatorname{tr}(L_{uu}(M_k)) \label{eq:quadratic_sdp_cost}\\
    \text{s.t.}\quad & L_x(M_1) = x_i, \quad L_{xx}(M_1) = x_ix_i^\top, \label{eq:sdp-initial-condition}\\
    & \|L_u (M_k)\|_2 \leq b_u, \label{eq:sdp_boud_cons} \\
    & L_x (M_{k+1}) = A_k L_x (M_k) + B_k L_u (M_k), \label{sdp-dynamics-constraints} \\
    & \eqref{quadratic-dynamics-sdp},\ \eqref{psd-M},\ \eqref{sdp-pc-constraint} , \ \eqref{lower-bound} ,\label{final-constraints}
    \end{alignat}
\end{subequations}
where the objective in \eqref{eq:quadratic_sdp_cost} still minimizes the squared L2 norm of the controls, resulting in a minimum-energy solution. Eq. \eqref{eq:sdp-initial-condition} enforces the initial state to match a user-defined initial condition $x_i$, while \eqref{eq:sdp_boud_cons} limits the control $L_2$ norm by the upper bound $b_u$. The linear dynamics, expressed in terms of the variable $M$, are captured by \eqref{sdp-dynamics-constraints}. Finally, \eqref{final-constraints} consolidates the remaining constraints: the redundant constraint for the dynamics of the outer-product terms, the positive-definite constraints on $M$, the PoC constraint, and the lower bound magnitude constraint (only in Example 2). We have reformulated the original nonconvex collision-avoidance problem as a convex semidefinite program (SDP) that can be solved to global optimality.

In cases where the primary satellite cannot achieve the desired PoC in the given time due to thrust limits, we modify \eqref{eq:sdp} to offer a contingency plan. This modification consists of relaxing the PoC constraint in \eqref{sdp-pc-constraint} by introducing a weighted penalty on the L1 norm between the desired and achieved probability of collision. The contingency plan objective becomes, 
\begin{equation}
J = \sum_{i=1}^{N-1} \operatorname{tr}(L_{uu}(M_k))  + 
\alpha|| \operatorname{tr}(C^{-1}b) + \operatorname{tr}(C^{-1}c(M_N)) - p||_1
\label{modified-objective}
\end{equation} 
where $\alpha$ is the weight on the collision risk term of the objective function. This updated optimization problem minimizes control effort along with the risk of collision.



In the following section, we solve both versions of the SDP and perform a rank check to determine whether the obtained solution corresponds to the global optimum of the original nonconvex formulation.

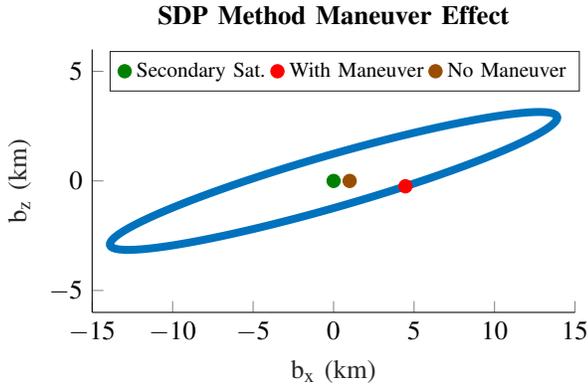
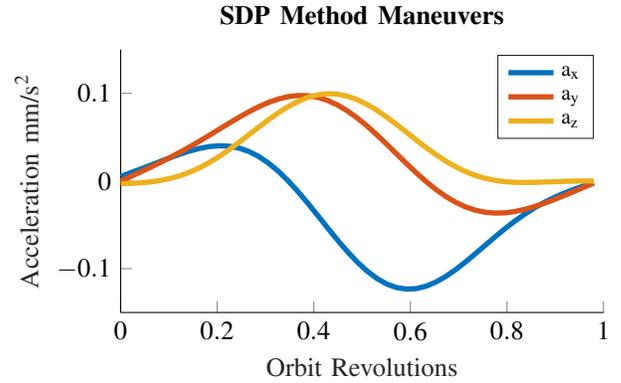
\begin{figure*}[t]
  \centering
  \begin{subfigure}[t]{0.45\textwidth}
    \centering
    \scalebox{1.0}{
%
%
\definecolor{mycolor1}{rgb}{0.00000,0.44700,0.74100}%
\begin{tikzpicture}

\begin{axis}[%
width= \columnwidth,
height=2.0in,
at={(0.0in,0.0in)},
xmin=-15,
xmax=15,
xlabel style={font=\color{white!15!black}},
xlabel={$\text{b}_\text{x}\text{ (km)}$},
ymin=-6,
ymax=6,
ylabel style={font=\color{white!15!black}},
ylabel={$\text{b}_\text{z}\text{ (km)}$},
axis background/.style={fill=white},
title style={font=\bfseries},
title={SDP Method Maneuver Effect},
axis x line*=bottom,
axis y line*=left,
legend style={at={(1.01,0.84)}, anchor=south east, legend cell align=left, align=left, draw=white!15!black, legend columns =3}
]
\addplot [color=mycolor1, line width=3.0pt, forget plot]
  table[row sep=crcr]{%
-0.249005170438108	1.18755986979187\\
0.633045618245353	1.37001044949471\\
1.51254735565365	1.54694448065409\\
2.38595859870489	1.71764951237428\\
3.24976242858836	1.88143817574184\\
4.10048061215531	2.03765095161757\\
4.93468760753619	2.18565882628733\\
5.74902435758877	2.3248658242783\\
6.54021181563566	2.45471140814181\\
7.30506414902799	2.57467273553983\\
8.04050156736906	2.68426676454653\\
8.74356272374325	2.78305219868763\\
9.41141663901492	2.87063126388554\\
10.0413741011821	2.9466513101552\\
10.6308984938843	3.01080623160109\\
11.177616010461	3.06283769899762\\
11.6793252124334	3.10253619998977\\
12.1340058939205	3.12974188272537\\
12.5398272162945	3.14434519952214\\
12.8951550803224	3.14628734797747\\
13.1985587061064	3.13556050774502\\
13.44881639433	3.11220787202442\\
13.6449204456096	3.07632347363755\\
13.7860812181444	3.02805180639154\\
13.8717303073251	2.96758724325321\\
13.9015228344991	2.89517325367773\\
13.8753388356743	2.81110142324305\\
13.7932837445727	2.7157102795377\\
13.6556879680853	2.60938392902967\\
13.4631055558403	2.49255051040523\\
13.2163119692412	2.36568047060557\\
12.9163009589569	2.22928467050295\\
12.5642805634388	2.08391232784436\\
12.1616682445763	1.93014880574541\\
11.7100851800779	1.76861325563971\\
11.2113497355609	1.59995612417458\\
10.6674701426349	1.42485653409194\\
10.0806364124616	1.24401954964076\\
9.45321151735308	1.05817333753211\\
8.78772187591631	0.868066234868912\\
8.08684718005689	0.674463735856462\\
7.35340960480571	0.478145409427492\\
6.59036244441608	0.279901760193201\\
5.80077822049013	0.0805310453601929\\
4.98783631001881	-0.119163939569549\\
4.15481014315278	-0.318379093372815\\
3.30505402225463	-0.516312246936985\\
2.44198961530708	-0.712166393311931\\
1.56909217806371	-0.9051528969757\\
0.689876560420579	-1.09449466939142\\
-0.192116946643726	-1.27942929806858\\
-1.07333686657038	-1.45921211652901\\
-1.95023483776291	-1.63311920281601\\
-2.81927990158701	-1.80045029447251\\
-3.67697272029469	-1.96053160825088\\
-4.51985966762235	-2.11271855320024\\
-5.3445467353248	-2.25639832620671\\
-6.14771319964857	-2.39099237953528\\
-6.92612499271441	-2.51595875043732\\
-7.67664772496726	-2.63079424344332\\
-8.39625930625665	-2.7350364565534\\
-9.08206211472713	-2.82826564316701\\
-9.73129466451885	-2.91010640225433\\
-10.3413427252964	-2.98022918996369\\
-10.9097498488316	-3.03835164657839\\
-11.4342272602538	-3.0842397334796\\
-11.9126630741382	-3.11770867553728\\
-12.3431307983233	-3.13862370513442\\
-12.7238970912146	-3.14690060482867\\
-13.0534287413398	-3.14250604646623\\
-13.3303988410499	-3.1254577253826\\
-13.5536921295084	-3.09582428914968\\
-13.7224094834525	-3.05372506115624\\
-13.8358715376457	-2.99932956013477\\
-13.8936214204414	-2.93285681756931\\
-13.8954265934441	-2.85457449573303\\
-13.8412797878592	-2.76479780990672\\
-13.7313990337621	-2.66388825911808\\
-13.5662267821677	-2.55225217051273\\
-13.3464281234368	-2.43033906321822\\
-13.0728881091914	-2.29863983828916\\
-12.7467081885247	-2.15768480202192\\
-12.369201772854	-2.00804153059834\\
-11.9418889472768	-1.85031258465679\\
-11.4664903497241	-1.68513308299313\\
-10.9449202425595	-1.51316814516147\\
-10.3792788045194	-1.33511021327251\\
-9.77184367403535	-1.15167626377354\\
-9.12506077798727	-0.963604920437306\\
-8.44153448281958	-0.771653480184829\\
-7.72401710767657	-0.576594863717932\\
-6.97539784178465	-0.37921450324041\\
-6.19869111070708	-0.180307179799734\\
-5.39702443831609	0.0193261770156987\\
-4.57362585335813	0.218881714137346\\
-3.73181089132146	0.417555891849097\\
-2.87496924394483	0.614548719358121\\
-2.00655111012532	0.809066976075432\\
-1.13005330318504	1.00032740563524\\
-0.249005170438112	1.18755986979187\\
};

\addplot[only marks, mark=*, mark options={}, mark size=2.5pt, color=black!50!green, fill=black!50!green] table[row sep=crcr]{%
x	y\\
0	0\\
};
\addlegendentry{Secondary Sat.}

\addplot[only marks, mark=*, mark options={}, mark size=2.5pt, color=red, fill=red] table[row sep=crcr]{%
x	y\\
4.46524923153157	-0.244754920468287\\
};
\addlegendentry{With Maneuver}

\addplot[only marks, mark=*, mark options={}, mark size=2.5pt, color=black!40!orange, fill=black!40!orange] table[row sep=crcr]{%
x	y\\
0.998461159101683	-6.60170702826424e-17\\
};
\addlegendentry{No Maneuver}

\end{axis}

\end{tikzpicture}%
}
    \caption{The maneuver end state in the b-plane.}
    \label{sdp-manuever-effect}
  \end{subfigure}
  \hfill
  \begin{subfigure}[t]{0.45\textwidth}
    \centering
    \scalebox{1.0}{
%
%
\definecolor{mycolor1}{rgb}{0.00000,0.44700,0.74100}%
\definecolor{mycolor2}{rgb}{0.85000,0.32500,0.09800}%
\definecolor{mycolor3}{rgb}{0.92900,0.69400,0.12500}%
\begin{tikzpicture}[scale=1.0]

\begin{axis}[%
width=\columnwidth,
height=2.0 in,
at={(0 in,0 in)},
xmin=0,
xmax=1,
xlabel style={font=\color{white!15!black}},
xlabel={Orbit Revolutions},
ymin=-0.15,
ymax=0.15,
ylabel style={font=\color{white!15!black}},
ylabel={$\text{Acceleration mm/s}^\text{2}$},
axis background/.style={fill=white},
title style={font=\bfseries},
title={SDP Method Maneuvers},
axis x line*=bottom,
axis y line*=left,
legend style={legend cell align=left, align=left, draw=white!15!black}
]
\addplot [color=mycolor1, line width=2.0pt]
  table[row sep=crcr]{%
0	0.00521701830120052\\
0.0204081632653061	0.00938172247476459\\
0.0408163265306122	0.0136062152117007\\
0.0612244897959184	0.0179071141421043\\
0.0816326530612245	0.022244432103286\\
0.102040816326531	0.0265195563706589\\
0.122448979591837	0.0305786117577079\\
0.142857142857143	0.0342208423449209\\
0.163265306122449	0.037210544878963\\
0.183673469387755	0.0392924274038263\\
0.204081632653061	0.0402091334021469\\
0.224489795918367	0.039719890495322\\
0.244897959183673	0.0376193064743889\\
0.26530612244898	0.0337549175081418\\
0.285714285714286	0.0280422907011698\\
0.306122448979592	0.0204764837267953\\
0.326530612244898	0.0111372406369591\\
0.346938775510204	0.000196268628683495\\
0.36734693877551	-0.0120904240691249\\
0.387755102040816	-0.0253927295441214\\
0.408163265306122	-0.0393174339505619\\
0.428571428571429	-0.0534281019394666\\
0.448979591836735	-0.0672651582839727\\
0.469387755102041	-0.0803689832543763\\
0.489795918367347	-0.092303417172687\\
0.510204081632653	-0.102677608884514\\
0.530612244897959	-0.111165269235916\\
0.551020408163265	-0.117520301532752\\
0.571428571428571	-0.121587632496568\\
0.591836734693878	-0.123308761736733\\
0.612244897959184	-0.122722281879643\\
0.63265306122449	-0.11995855360645\\
0.653061224489796	-0.115229735992535\\
0.673469387755102	-0.108815569877159\\
0.693877551020408	-0.101045946169239\\
0.714285714285714	-0.0922810944901191\\
0.73469387755102	-0.0828909238179359\\
0.755102040816327	-0.0732343637293391\\
0.775510204081633	-0.0636402905935241\\
0.795918367346939	-0.0543912620793997\\
0.816326530612245	-0.0457106330291203\\
0.836734693877551	-0.0377537646571133\\
0.857142857142857	-0.0306050570509284\\
0.877551020408163	-0.0242800329122156\\
0.897959183673469	-0.0187304917842914\\
0.918367346938776	-0.0138574863686851\\
0.938775510204082	-0.00952240932287366\\
0.959183673469388	-0.0055672509075185\\
0.979591836734694	-0.00183299304401723\\
};
\addlegendentry{$\text{a}_\text{x}$}

\addplot [color=mycolor2, line width=2.0pt]
  table[row sep=crcr]{%
0	-6.55079639108703e-05\\
0.0204081632653061	0.0050273590297869\\
0.0408163265306122	0.0102028511866663\\
0.0612244897959184	0.0155057564687084\\
0.0816326530612245	0.020990271043601\\
0.102040816326531	0.0267060394450336\\
0.122448979591837	0.0326883117839742\\
0.142857142857143	0.0389500158519198\\
0.163265306122449	0.0454747402799446\\
0.183673469387755	0.052210718686497\\
0.204081632653061	0.0590682175931007\\
0.224489795918367	0.0659195159427817\\
0.244897959183673	0.0726018631950733\\
0.26530612244898	0.0789232302703862\\
0.285714285714286	0.0846707733088043\\
0.306122448979592	0.089621531578433\\
0.326530612244898	0.0935539983004001\\
0.346938775510204	0.0962617352688588\\
0.36734693877551	0.0975667896940112\\
0.387755102040816	0.0973303411789779\\
0.408163265306122	0.0954640859292663\\
0.428571428571429	0.0919374073462789\\
0.448979591836735	0.0867824509097507\\
0.469387755102041	0.0800953902231525\\
0.489795918367347	0.0720339318497931\\
0.510204081632653	0.0628114933050791\\
0.530612244897959	0.0526882361409165\\
0.551020408163265	0.0419595091546081\\
0.571428571428571	0.0309424846225591\\
0.591836734693878	0.0199617696714403\\
0.612244897959184	0.0093347480775627\\
0.63265306122449	-0.000642388857997606\\
0.653061224489796	-0.00970721621295969\\
0.673469387755102	-0.0176416138724965\\
0.693877551020408	-0.0242798958800196\\
0.714285714285714	-0.029513867508558\\
0.73469387755102	-0.0332945056478048\\
0.755102040816327	-0.0356303220429701\\
0.775510204081633	-0.0365827998150294\\
0.795918367346939	-0.036259500862824\\
0.816326530612245	-0.034804040718223\\
0.836734693877551	-0.0323835534785249\\
0.857142857142857	-0.0291782547946068\\
0.877551020408163	-0.025369588192633\\
0.897959183673469	-0.0211307706727989\\
0.918367346938776	-0.0166082602217147\\
0.938775510204082	-0.0119245511808353\\
0.959183673469388	-0.00716739574194598\\
0.979591836734694	-0.0023890419718669\\
};
\addlegendentry{$\text{a}_\text{y}$}

\addplot [color=mycolor3, line width=2.0pt]
  table[row sep=crcr]{%
0	-0.00305469026193876\\
0.0204081632653061	-0.00268367536898501\\
0.0408163265306122	-0.00206371134964698\\
0.0612244897959184	-0.00104156434720853\\
0.0816326530612245	0.000541933988515989\\
0.102040816326531	0.00283784300113055\\
0.122448979591837	0.00597412291339858\\
0.142857142857143	0.0100433492169855\\
0.163265306122449	0.0150917788187357\\
0.183673469387755	0.0211112136745526\\
0.204081632653061	0.0280340853915609\\
0.224489795918367	0.0357324927272095\\
0.244897959183673	0.0440209780837723\\
0.26530612244898	0.0526631594103952\\
0.285714285714286	0.0613820285742078\\
0.306122448979592	0.069873560893528\\
0.326530612244898	0.0778204940394838\\
0.346938775510204	0.0849124667195806\\
0.36734693877551	0.0908625028824312\\
0.387755102040816	0.0954204583795512\\
0.408163265306122	0.098390559124527\\
0.428571428571429	0.0996411948139058\\
0.448979591836735	0.0991139211377284\\
0.469387755102041	0.096826971494797\\
0.489795918367347	0.0928739491724377\\
0.510204081632653	0.087418519054304\\
0.530612244897959	0.0806850397658407\\
0.551020408163265	0.0729458668146662\\
0.571428571428571	0.0645062195294639\\
0.591836734693878	0.0556875167993631\\
0.612244897959184	0.0468102754050527\\
0.63265306122449	0.0381774840362815\\
0.653061224489796	0.0300595798744945\\
0.673469387755102	0.0226818269433812\\
0.693877551020408	0.0162147986502045\\
0.714285714285714	0.0107682980745247\\
0.73469387755102	0.00638932903640905\\
0.755102040816327	0.00306403316007083\\
0.775510204081633	0.000723304118071064\\
0.795918367346939	-0.000748098142985401\\
0.816326530612245	-0.00149963716279277\\
0.836734693877551	-0.00170191265312873\\
0.857142857142857	-0.00153169625524626\\
0.877551020408163	-0.00115745678335316\\
0.897959183673469	-0.000723157857870396\\
0.918367346938776	-0.000340959901928292\\
0.938775510204082	-8.13589706922631e-05\\
0.959183673469388	3.29882294049571e-05\\
0.979591836734694	2.68367140548465e-05\\
};
\addlegendentry{$\text{a}_\text{z}$}

\end{axis}

\end{tikzpicture}%
}
    \caption{Low-thrust collision avoidance maneuvers, represented as additive accelerations. }
    \label{sdp-manevuers}
  \end{subfigure}
  \caption{ (Example 1) The boundary of the ellipse in (a) represents all the relative positions that have a $P_c =1 \times 10^{-6}$. With no maneuvers, the relative position of the primary spacecraft is represented by the brown marker. After the maneuvers in (b) are applied, the relative position is represented by the red marker, showing the success of the maneuvers to achieve the desired $P_c$ value.}
  \label{sdp-results}
\end{figure*}
\section{EXPERIMENTS AND RESULTS}
\label{experiments-results}


To evaluate our approach, we conduct a set of case studies using a simulated CDM and a high-fidelity simulation in LEO. The prediction horizon is one orbital revolution, which is representative of a rapid response to a future conjunction event. All the simulation parameters --- including orbital elements, initial conditions, and force models --- are provided in our open-source implementation available on Github\footnote{\href{https://github.com/RoboticExplorationLab/cvx-spacecraft-cola.git}{https://github.com/RoboticExplorationLab/cvx-spacecraft-cola.git}}. The hard-body radius was set to 10$m$. We solve our semidefinite program using MOSEK \cite{andersen2000mosek} and compare to other existing methods of solving the collision-avoidance problem. Examples 1 and 3 do not include the lower bound constraint while Example 2 does. In all examples, the state and uncertainty of the secondary spacecraft is known at the TCA from the simulated CDM. The secondary satellite is also  located at the origin of the b-plane frame, as the frame is defined to be centered on the secondary spacecraft.

\subsection{Example 1}
The target post-maneuver PoC is set to $P_c = 1 \times 10^{-6}$. The resulting continuous low-thrust maneuvers from the SDP are shown in Fig. \ref{sdp-manevuers}. These maneuvers remain within the set acceleration limits and exhibit the expected smoothness due to the chosen cost function. Initially, the probability of collision is $1 \times 10^{-5}$, corresponding to a relative position inside the $P_c =1 \times 10^{-6}$ ellipse on the b-plane. After applying the optimized maneuvers, the final relative position lies exactly on the boundary of the target $P_c$ ellipse, demonstrating that the collision probability has been reduced to the desired level.

\subsubsection{Comparison to Other Methods}


\begin{figure}[t]
    \centering
    \scalebox{1.0}{\input{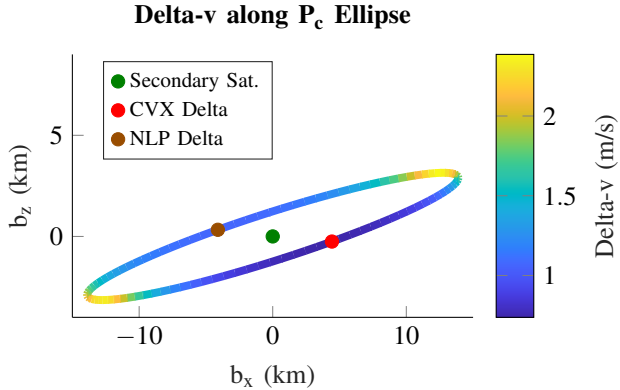}}
    \caption{The solution obtained from successive linearizations, shown by the red marker, closely approximates the global optimum after 100 iterations around the ellipse. The heat map illustrates the variation in delta-v as a function of the linearization point on the ellipse. In contrast, the nonlinear trajectory optimization solution, shown by the brown marker,  converges to a local minimum, which is suboptimal.
    }
    \label{deltav-heatmap-combined}
\end{figure}

\begin{figure}[t]
    \centering
    \scalebox{1.0}{
%
%
\definecolor{mycolor1}{rgb}{0.00000,0.44700,0.74100}%
\definecolor{mycolor2}{rgb}{0.85000,0.32500,0.09800}%
\definecolor{mycolor3}{rgb}{0.92900,0.69400,0.12500}%
\begin{tikzpicture}


\begin{axis}[%
width=\columnwidth,
height=2.0in,
at={(0.758in,0.481in)},
xmin=0,
xmax=50,
xlabel style={font=\color{white!15!black}},
xlabel={Knot Point},
ymode=log,
ymin=5000,
ymax=10000000000,
yminorticks=true,
ylabel style={font=\color{white!15!black}},
ylabel={Eigenvalue Ratio},
axis background/.style={fill=white},
title style={font=\bfseries},
title={Tightness Check},
axis x line*=bottom,
axis y line*=left,
legend style={at={(0.95,0.8)}, anchor=south east, legend cell align=left, align=left, draw=white!15!black, legend columns =3}
]
\addplot [color=mycolor1, line width=2.0pt]
  table[row sep=crcr]{%
1	1005468.75874622\\
2	1665134.88381084\\
3	2392721.3981067\\
4	3001271.08207353\\
5	3292347.35016675\\
6	3850199.79482864\\
7	3922134.95143741\\
8	4019592.09384384\\
9	3738911.97482491\\
10	3345114.54577276\\
11	2865019.56009445\\
12	2441755.41242648\\
13	2090861.07401908\\
14	1869682.33222359\\
15	1692814.53794787\\
16	1542683.13561562\\
17	1338281.856707\\
18	1136479.06607318\\
19	937368.580776452\\
20	767918.829223493\\
21	618958.506739593\\
22	498225.890954085\\
23	409680.510427898\\
24	356556.607454267\\
25	338791.577062785\\
26	360879.647689101\\
27	433035.700031431\\
28	574546.815733644\\
29	813612.523425059\\
30	1191578.4325963\\
31	1758629.34754228\\
32	2575285.43874837\\
33	3707219.20954466\\
34	5226914.74998942\\
35	7207886.38557989\\
36	9723106.06106202\\
37	12887508.560051\\
38	16799443.3179362\\
39	21576776.2788468\\
40	27375336.7115133\\
41	34310660.2064056\\
42	42734438.956998\\
43	52770396.0125396\\
44	64918683.1472966\\
45	79571369.5115448\\
46	97272555.6497406\\
47	118057544.908813\\
48	142019454.243038\\
49	169098835.792764\\
50	201236327.132681\\
};
\addlegendentry{Example 1}

\addplot [color=mycolor2, line width=2.0pt]
  table[row sep=crcr]{%
1	982457.07279108\\
2	1568490.28226186\\
3	3092166.59062602\\
4	4276946.6446265\\
5	5068592.5493718\\
6	5363206.31351124\\
7	4615656.36787741\\
8	3686838.41675115\\
9	2856067.89204367\\
10	2221063.72796969\\
11	1755959.20135363\\
12	1426482.78211544\\
13	1173362.34699422\\
14	964604.255983482\\
15	803493.425750402\\
16	669869.159226079\\
17	555847.972645746\\
18	457849.922565898\\
19	372391.564196658\\
20	298285.673873718\\
21	234821.968179993\\
22	182023.932431933\\
23	139651.608697286\\
24	107313.65788561\\
25	84328.287460032\\
26	69882.5545946071\\
27	63326.0053525752\\
28	64646.3352573696\\
29	74970.7617286486\\
30	97026.8690122159\\
31	135445.500205131\\
32	196831.642211532\\
33	289691.257123711\\
34	424274.055258811\\
35	612351.871791878\\
36	867186.949690773\\
37	1203475.49638632\\
38	1638075.24895987\\
39	2189231.72958718\\
40	2878142.81048538\\
41	3729275.78742435\\
42	4771366.8752961\\
43	6038386.56093332\\
44	7572487.90381723\\
45	9420567.88771498\\
46	11652576.6644439\\
47	14348613.9222957\\
48	17609314.0872818\\
49	21556056.8381737\\
50	26325119.092605\\
};
\addlegendentry{Example 2}

\addplot [color=mycolor3, line width=2.0pt]
  table[row sep=crcr]{%
1	428183.835889065\\
2	658178.956344628\\
3	770369.835918349\\
4	810884.256063752\\
5	815920.666507458\\
6	805603.533397425\\
7	792379.411986003\\
8	691993.656640569\\
9	565068.464956999\\
10	458593.225123684\\
11	375891.19666317\\
12	313650.345804331\\
13	263267.041873522\\
14	218904.988255548\\
15	185159.306678223\\
16	155232.194206151\\
17	129698.309810284\\
18	107306.740024191\\
19	87418.373041743\\
20	70007.9716021529\\
21	54975.1370734749\\
22	42314.1336114759\\
23	31974.795289809\\
24	23885.6830786892\\
25	17853.3143728533\\
26	13631.2713189844\\
27	10957.194827223\\
28	9607.34135053899\\
29	9466.25608856472\\
30	10590.5493762277\\
31	13256.8279213808\\
32	17992.1938852394\\
33	25587.8153246367\\
34	37097.6658822486\\
35	53824.5693697659\\
36	77273.5496103526\\
37	109148.255519269\\
38	151735.058601743\\
39	206901.490617991\\
40	276724.31868005\\
41	363606.952122763\\
42	470223.229287932\\
43	599431.350417572\\
44	754697.119932223\\
45	940139.694608247\\
46	1160677.02491107\\
47	1422349.5813619\\
48	1732606.12647106\\
49	2100814.23119352\\
50	2538827.86742026\\
};
\addlegendentry{Example 3}
\end{axis}

\end{tikzpicture}%
}
    \caption{The ratio between the largest and second-largest eigenvalues in each moment matrix is above $1 \times 10^4$, ensuring a rank 1 moment matrix and tight relaxation. For Example 3, only the case where delta-v limit is set to 0.004 $m/s$ is plotted. The other cases had a similar tightness result.}
    \label{tightness-combined}
\end{figure}
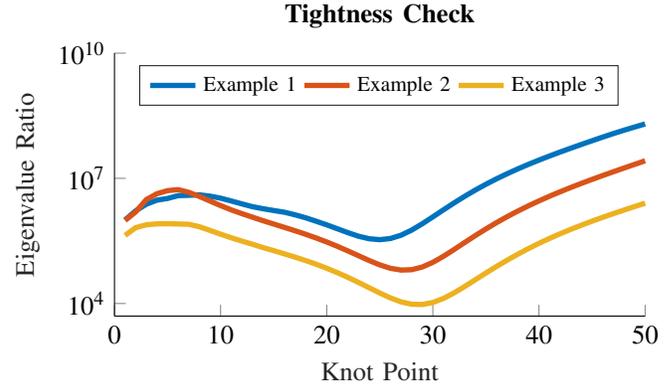

For comparison, we evaluate two alternative methods commonly used in spacecraft collision avoidance: (i) solving the nonlinear program directly using a general purpose solver, and (ii) a collision avoidance half-plane approximation method. 

We solve the NLP \eqref{eq:nlp} using the IPOPT solver \cite{wachter2006implementation}. This approach requires a good initial guess for both the state and control trajectories, which is non-trivial to obtain. In our implementation, we initialize the state trajectory with the unperturbed reference trajectory and set all the control inputs to zero. With this initialization, IPOPT converges to a local minimum with higher cost than the global minimum (Fig. \ref{deltav-heatmap-combined}). 

The second method samples points along the $P_c$ ellipse in the b-plane. At each sample point, the local tangent is computed, and the nonlinear ellipse constraint is conservatively replaced with a linear half-plane constraint. This approach follows the validation procedure for the SCP algorithm implemented in \cite{armellin2021collision}. We solve 100 of these convex problems with different linearizations and select the solution with the lowest cost. As the number of samples increases, the union of these half-planes converges to the true nonconvex constraint, allowing this method to closely match the globally optimal solution. However, its performance depends on the quality of the half-plane approximation, and solving a large number of problems may become computationally expensive. Figure \ref{deltav-heatmap-combined} shows the optimal solution from the sampling approach. The NLP solution is also plotted, and we can see that it converges to a local solution with 108\% percent greater cost than the global solution obtained from the SDP.

\subsubsection{Tightness}
To assess the tightness of the convex relaxation, we compute the ratio between the largest and second-largest eigenvalues of the moment matrices at each point along the trajectory. In all cases, the ratios exceed $10^4$ which, given floating-point precision, indicates that the moment matrices are effectively rank 1, and that the convex relaxation remains tight. This ratio is depicted in Fig \ref{tightness-combined}.

\subsection{Example 2}
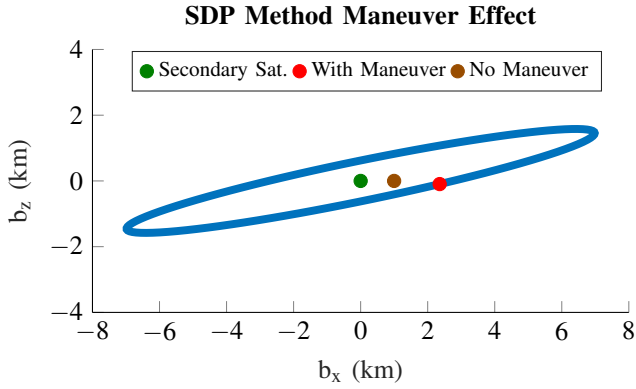
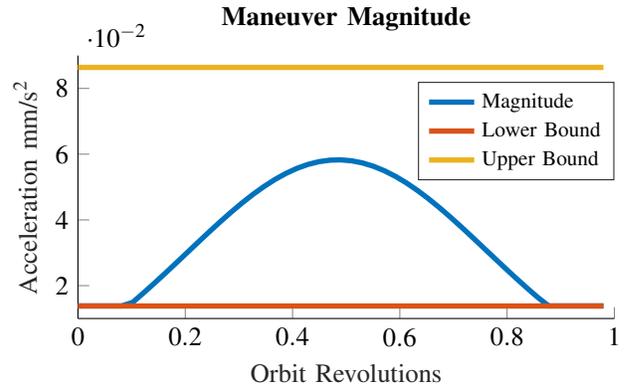
\begin{figure*}[t]
  \centering
  \begin{subfigure}[t]{0.49\textwidth}
    \centering
    \scalebox{1.0}{
%
%
\definecolor{mycolor1}{rgb}{0.00000,0.44700,0.74100}%
\begin{tikzpicture}

\begin{axis}[%
width= \columnwidth,
height=2.0in,
at={(0.0in,0.0in)},
xmin=-8,
xmax=8,
xlabel style={font=\color{white!15!black}},
xlabel={$\text{b}_\text{x}\text{ (km)}$},
ymin=-4,
ymax=4,
ylabel style={font=\color{white!15!black}},
ylabel={$\text{b}_\text{z}\text{ (km)}$},
axis background/.style={fill=white},
title style={font=\bfseries},
title={SDP Method Maneuver Effect},
axis x line*=bottom,
axis y line*=left,
legend style={at={(0.95,0.84)}, anchor=south east, legend cell align=left, align=left, draw=white!15!black, legend columns =3}
]
\addplot [color=mycolor1, line width=3.0pt]
  table[row sep=crcr, forget plot]{%
-0.125088445051196	0.596574027985453\\
0.318012240040882	0.688228587903076\\
0.759832402082754	0.777111894203204\\
1.19859298721955	0.862866045182178\\
1.63252726143274	0.945145739140396\\
2.05988792455446	1.0236196647904\\
2.47895414602862	1.09797183533047\\
2.88803849408813	1.16790286081198\\
3.28549373044714	1.23313115367704\\
3.66971944314818	1.2933940626122\\
4.03916849085627	1.34844893015259\\
4.39235323265103	1.39807406977791\\
4.72785151823162	1.44206965856576\\
5.04431241441401	1.48025854180807\\
5.3404616448622	1.51248694635055\\
5.61510672114918	1.53862509978292\\
5.86714174448703	1.5585677529866\\
6.09555185879094	1.57223460393578\\
6.29941733714656	1.57957062104532\\
6.47791728522557	1.58054626476351\\
6.63033294673753	1.57515760651746\\
6.75605059760767	1.56342634453203\\
6.8545640172272	1.54539971645875\\
6.92547652682502	1.5211503091665\\
6.9685025867531	1.49077576645975\\
6.9834689462539	1.45439839590148\\
6.97031534107997	1.41216467632378\\
6.92909473615688	1.36424466800924\\
6.8599731123121	1.31083132791828\\
6.76322879792881	1.25213973271954\\
6.63925134821571	1.18840621275206\\
6.48853997660575	1.11988740040639\\
6.31170154459974	1.04685919675657\\
6.10944811814936	0.969615660603833\\
5.88259410041884	0.888467824405654\\
5.63205295247087	0.803742441857777\\
5.35883351508129	0.715780672172424\\
5.06403594649337	0.624936706350562\\
4.74884729246886	0.531576340979741\\
4.41453670647364	0.436075505300354\\
4.06245033924466	0.338818747471253\\
3.69400591831609	0.240197686130051\\
3.31068703933082	0.140609433483081\\
2.91403719212453	0.0404549962746966\\
2.50565354563683	-0.0598623389253461\\
2.08718051667565	-0.159938629614576\\
1.66030314843118	-0.259370903891691\\
1.22674032540162	-0.357758783082179\\
0.788237852052031	-0.454706093921927\\
0.34656142307586	-0.549822463807135\\
-0.0965104864343391	-0.642724892687004\\
-0.539193782278445	-0.733039295269768\\
-0.979705935066399	-0.820402007332132\\
-1.41627315783495	-0.904461250066747\\
-1.84713754846168	-0.984878546571304\\
-2.27056416811645	-1.06133008477557\\
-2.68484802724768	-1.13350802131828\\
-3.08832095097357	-1.20112172112373\\
-3.47935829623352	-1.26389892768665\\
-3.85638549365238	-1.32158685935302\\
-4.21788438777552	-1.37395322718262\\
-4.56239935014511	-1.42078717029459\\
-4.88854314060208	-1.46190010492977\\
-5.19500249321268	-1.49712648381092\\
-5.48054340432688	-1.52632446274323\\
-5.7440161014755	-1.54937647177078\\
-5.98435967309817	-1.56618968858929\\
-6.20060634045972	-1.57669641230873\\
-6.39188535455362	-1.58085433606098\\
-6.55742650230084	-1.5786467173546\\
-6.69656320792611	-1.57008244549096\\
-6.80873521702332	-1.55519600577014\\
-6.89349085250225	-1.53404734063082\\
-6.95048883333279	-1.50672160828323\\
-6.97949964876323	-1.47332883980709\\
-6.980406482479	-1.43400349609528\\
-6.9532056829808	-1.38890392642729\\
-6.89800677828786	-1.33821173085256\\
-6.81503203490727	-1.28213102895119\\
-6.70461556284519	-1.22088763791638\\
-6.56720197026382	-1.15472816326837\\
-6.40334457320121	-1.08391900586099\\
-6.21370316756289	-1.00874528917946\\
-5.9990413723566	-0.929509711248775\\
-5.76022355486811	-0.84653132577555\\
-5.49821135015917	-0.760144257431383\\
-5.21405978890258	-0.670696356450717\\
-4.90891304914599	-0.578547797960726\\
-4.58399984911076	-0.484069631683187\\
-4.24062849957735	-0.38764228784822\\
-3.88018163577953	-0.289654045336017\\
-3.50411065002037	-0.190499468214838\\
-3.1139298474276	-0.0905778169707575\\
-2.71121034838137	0.00970855917338801\\
-2.297573762167	0.109955842376319\\
-1.87468565732673	0.209760372210229\\
-1.44424885500303	0.308720271058864\\
-1.00799657227887	0.406437062336842\\
-0.567685443124241	0.502517275014219\\
-0.125088445051198	0.596574027985453\\
};

\addplot[only marks, mark=*, mark options={}, mark size=2.5pt, color=black!50!green, fill=black!50!green] table[row sep=crcr]{%
x	y\\
0	0\\
};
\addlegendentry{Secondary Sat.}

\addplot[only marks, mark=*, mark options={}, mark size=2.5pt, color=red, fill=red] table[row sep=crcr]{%
x	y\\
2.35728799114708	-0.0956482523775415\\
};
\addlegendentry{With Maneuver}

\addplot[only marks, mark=*, mark options={}, mark size=2.5pt, color=black!40!orange, fill=black!40!orange] table[row sep=crcr]{%
x	y\\
0.998461159101683	-6.60170702826424e-17\\
};
\addlegendentry{No Maneuver}

\end{axis}

\end{tikzpicture}%
}
    \caption{The maneuver end state in the b-plane.}
    \label{sdp-manuever-effect-example-2}
  \end{subfigure}
  \hfill
  \begin{subfigure}[t]{0.49\textwidth}
    \centering
    \scalebox{1.0}{
%
%
\definecolor{mycolor1}{rgb}{0.00000,0.44700,0.74100}%
\definecolor{mycolor2}{rgb}{0.85000,0.32500,0.09800}%
\definecolor{mycolor3}{rgb}{0.92900,0.69400,0.12500}%
\begin{tikzpicture}

\begin{axis}[%
width=\columnwidth,
height=2.0in,
at={(0.0in,0.0in)},
xmin=0,
xmax=1,
xlabel style={font=\color{white!15!black}},
xlabel={Orbit Revolutions},
ymin=0.01,
ymax=0.09,
ylabel style={font=\color{white!15!black}},
ylabel={$\text{Acceleration mm/s}^\text{2}$},
axis background/.style={fill=white},
title style={font=\bfseries},
title={Maneuver Magnitude},
axis x line*=bottom,
axis y line*=left,
legend style={at={(1.0,0.90)}, legend cell align=left, align=left, draw=white!15!black}
]
\addplot [color=mycolor1, line width=2.0pt]
  table[row sep=crcr]{%
0	0.013830974581804\\
0.0204081632653061	0.013831278433528\\
0.0408163265306122	0.01383147772505\\
0.0612244897959184	0.0138315786588545\\
0.0816326530612245	0.0138316501688659\\
0.102040816326531	0.0149196286262717\\
0.122448979591837	0.0178156518476206\\
0.142857142857143	0.0208038178202091\\
0.163265306122449	0.0238685764071464\\
0.183673469387755	0.0269892463773573\\
0.204081632653061	0.0301400390258442\\
0.224489795918367	0.0332907438109975\\
0.244897959183673	0.036407574990288\\
0.26530612244898	0.0394541872306318\\
0.285714285714286	0.0423926627753044\\
0.306122448979592	0.0451845294469396\\
0.326530612244898	0.047791777418949\\
0.346938775510204	0.0501777306095642\\
0.36734693877551	0.0523079747204992\\
0.387755102040816	0.0541511910366219\\
0.408163265306122	0.0556799046655342\\
0.428571428571429	0.0568711130707307\\
0.448979591836735	0.0577067870260727\\
0.469387755102041	0.0581742476344368\\
0.489795918367347	0.0582664300911569\\
0.510204081632653	0.0579820166267774\\
0.530612244897959	0.0573254322379642\\
0.551020408163265	0.0563067098051738\\
0.571428571428571	0.0549412494545097\\
0.591836734693878	0.0532494705837526\\
0.612244897959184	0.0512564298196656\\
0.63265306122449	0.0489913105617695\\
0.653061224489796	0.0464868224261814\\
0.673469387755102	0.043778522518027\\
0.693877551020408	0.0409040892731512\\
0.714285714285714	0.0379024998215189\\
0.73469387755102	0.034813139568709\\
0.755102040816327	0.0316748353217554\\
0.775510204081633	0.0285248033146954\\
0.795918367346939	0.0253974830312568\\
0.816326530612245	0.0223233432500786\\
0.836734693877551	0.0193277552721281\\
0.857142857142857	0.0164298543867442\\
0.877551020408163	0.0138326422520836\\
0.897959183673469	0.0138316979590923\\
0.918367346938776	0.0138316681768721\\
0.938775510204082	0.0138316531060458\\
0.959183673469388	0.0138316276714756\\
0.979591836734694	0.0138315013718141\\
};
\addlegendentry{Magnitude}

\addplot [color=mycolor2, line width=2.0pt]
  table[row sep=crcr]{%
0	0.0138\\
0.0204081632653061	0.0138\\
0.0408163265306122	0.0138\\
0.0612244897959184	0.0138\\
0.0816326530612245	0.0138\\
0.102040816326531	0.0138\\
0.122448979591837	0.0138\\
0.142857142857143	0.0138\\
0.163265306122449	0.0138\\
0.183673469387755	0.0138\\
0.204081632653061	0.0138\\
0.224489795918367	0.0138\\
0.244897959183673	0.0138\\
0.26530612244898	0.0138\\
0.285714285714286	0.0138\\
0.306122448979592	0.0138\\
0.326530612244898	0.0138\\
0.346938775510204	0.0138\\
0.36734693877551	0.0138\\
0.387755102040816	0.0138\\
0.408163265306122	0.0138\\
0.428571428571429	0.0138\\
0.448979591836735	0.0138\\
0.469387755102041	0.0138\\
0.489795918367347	0.0138\\
0.510204081632653	0.0138\\
0.530612244897959	0.0138\\
0.551020408163265	0.0138\\
0.571428571428571	0.0138\\
0.591836734693878	0.0138\\
0.612244897959184	0.0138\\
0.63265306122449	0.0138\\
0.653061224489796	0.0138\\
0.673469387755102	0.0138\\
0.693877551020408	0.0138\\
0.714285714285714	0.0138\\
0.73469387755102	0.0138\\
0.755102040816327	0.0138\\
0.775510204081633	0.0138\\
0.795918367346939	0.0138\\
0.816326530612245	0.0138\\
0.836734693877551	0.0138\\
0.857142857142857	0.0138\\
0.877551020408163	0.0138\\
0.897959183673469	0.0138\\
0.918367346938776	0.0138\\
0.938775510204082	0.0138\\
0.959183673469388	0.0138\\
0.979591836734694	0.0138\\
};
\addlegendentry{Lower Bound}

\addplot [color=mycolor3, line width=2.0pt]
  table[row sep=crcr]{%
0	0.0864\\
0.0204081632653061	0.0864\\
0.0408163265306122	0.0864\\
0.0612244897959184	0.0864\\
0.0816326530612245	0.0864\\
0.102040816326531	0.0864\\
0.122448979591837	0.0864\\
0.142857142857143	0.0864\\
0.163265306122449	0.0864\\
0.183673469387755	0.0864\\
0.204081632653061	0.0864\\
0.224489795918367	0.0864\\
0.244897959183673	0.0864\\
0.26530612244898	0.0864\\
0.285714285714286	0.0864\\
0.306122448979592	0.0864\\
0.326530612244898	0.0864\\
0.346938775510204	0.0864\\
0.36734693877551	0.0864\\
0.387755102040816	0.0864\\
0.408163265306122	0.0864\\
0.428571428571429	0.0864\\
0.448979591836735	0.0864\\
0.469387755102041	0.0864\\
0.489795918367347	0.0864\\
0.510204081632653	0.0864\\
0.530612244897959	0.0864\\
0.551020408163265	0.0864\\
0.571428571428571	0.0864\\
0.591836734693878	0.0864\\
0.612244897959184	0.0864\\
0.63265306122449	0.0864\\
0.653061224489796	0.0864\\
0.673469387755102	0.0864\\
0.693877551020408	0.0864\\
0.714285714285714	0.0864\\
0.73469387755102	0.0864\\
0.755102040816327	0.0864\\
0.775510204081633	0.0864\\
0.795918367346939	0.0864\\
0.816326530612245	0.0864\\
0.836734693877551	0.0864\\
0.857142857142857	0.0864\\
0.877551020408163	0.0864\\
0.897959183673469	0.0864\\
0.918367346938776	0.0864\\
0.938775510204082	0.0864\\
0.959183673469388	0.0864\\
0.979591836734694	0.0864\\
};
\addlegendentry{Upper Bound}

\end{axis}

\end{tikzpicture}%
}
    \caption{Magnitude of the collision avoidance maneuvers}
    \label{sdp-maneuver-mag}
  \end{subfigure}
  \caption{(Example 2) The boundary of the ellipse in (a) represents all relative positions corresponding to a collision probability of $P_c =8 \times 10^{-6}$. The computed maneuvers in (b) satisfy the specified upper and lower bounds and, when applied, achieve the target  $P_c$ value.}
  \label{sdp-example2-results}
\end{figure*}

For this next example, we incorporate the lower-bound constraint \eqref{lower-bound} in the convex problem \eqref{eq:sdp}. The desired probability of collision was set to $P_c=8 \times 10^{-6}$ , with an upper bound of $8.64 \times 10^{-2} \ mm/s^2$ and the lower bound $1.38 \times 10^{-2} \ mm/s^2$ on the L2 norm of the control inputs. Solving this problem as an NLP with IPOPT reached the maximum number of iterations ($1000$) when initialized with the reference trajectory and zero controls. This highlights the difficulty of obtaining a good initial guess to converge to a local solution. 
In contrast, our SDP formulation was tight as shown in Fig. \ref{tightness-combined} and produced continuous maneuvers that satisfy the specified acceleration limits while reducing the probability of collision. The resulting change in the b-plane and maneuver magnitude plots are shown in Fig. \ref{sdp-example2-results}.

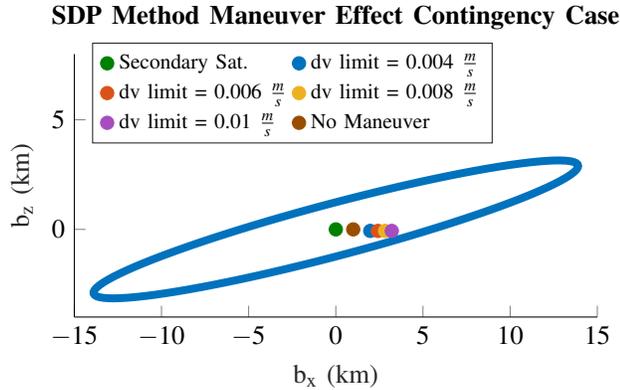
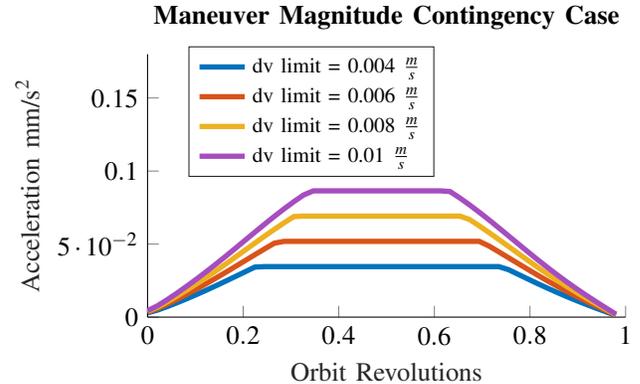
\begin{figure*}[t]
  \centering
  \begin{subfigure}[t]{0.48\textwidth}
    \centering
    \scalebox{1.0}{
%
%
\definecolor{mycolor1}{rgb}{0.00000,0.44700,0.74100}%

\definecolor{mycolor_1}{rgb}{0.00000,0.44700,0.74100}%
\definecolor{mycolor_2}{rgb}{0.85000,0.32500,0.09800}%
\definecolor{mycolor_3}{rgb}{0.92900,0.69400,0.12500}%
\definecolor{mycolor_4}{rgb}{0.65, 0.3, 0.75}%

\begin{tikzpicture}

\begin{axis}[%
width=\columnwidth,
height=2.0in,
at={(0.758in,0.516in)},
xmin=-15,
xmax=15,
xlabel style={font=\color{white!15!black}},
xlabel={$\text{b}_\text{x}\text{ (km)}$},
ymin=-4,
ymax=8,
ylabel style={font=\color{white!15!black}},
ylabel={$\text{b}_\text{z}\text{ (km)}$},
axis background/.style={fill=white},
title style={font=\bfseries},
title={SDP Method Maneuver Effect Contingency Case},
axis x line*=bottom,
axis y line*=left,
legend style={at={(0.8,0.66)}, anchor=south east, legend cell align=left, align=left, draw=white!15!black, legend columns =2}
]
\addplot [color=mycolor1, line width=3.0pt, forget plot]
  table[row sep=crcr]{%
-0.249005170438108	1.18755986979187\\
0.633045618245353	1.37001044949471\\
1.51254735565365	1.54694448065409\\
2.38595859870489	1.71764951237428\\
3.24976242858836	1.88143817574184\\
4.10048061215531	2.03765095161757\\
4.93468760753619	2.18565882628733\\
5.74902435758877	2.3248658242783\\
6.54021181563566	2.45471140814181\\
7.30506414902799	2.57467273553983\\
8.04050156736906	2.68426676454653\\
8.74356272374325	2.78305219868763\\
9.41141663901492	2.87063126388554\\
10.0413741011821	2.9466513101552\\
10.6308984938843	3.01080623160109\\
11.177616010461	3.06283769899762\\
11.6793252124334	3.10253619998977\\
12.1340058939205	3.12974188272537\\
12.5398272162945	3.14434519952214\\
12.8951550803224	3.14628734797747\\
13.1985587061064	3.13556050774502\\
13.44881639433	3.11220787202442\\
13.6449204456096	3.07632347363755\\
13.7860812181444	3.02805180639154\\
13.8717303073251	2.96758724325321\\
13.9015228344991	2.89517325367773\\
13.8753388356743	2.81110142324305\\
13.7932837445727	2.7157102795377\\
13.6556879680853	2.60938392902967\\
13.4631055558403	2.49255051040523\\
13.2163119692412	2.36568047060557\\
12.9163009589569	2.22928467050295\\
12.5642805634388	2.08391232784436\\
12.1616682445763	1.93014880574541\\
11.7100851800779	1.76861325563971\\
11.2113497355609	1.59995612417458\\
10.6674701426349	1.42485653409194\\
10.0806364124616	1.24401954964076\\
9.45321151735308	1.05817333753211\\
8.78772187591631	0.868066234868912\\
8.08684718005689	0.674463735856462\\
7.35340960480571	0.478145409427492\\
6.59036244441608	0.279901760193201\\
5.80077822049013	0.0805310453601929\\
4.98783631001881	-0.119163939569549\\
4.15481014315278	-0.318379093372815\\
3.30505402225463	-0.516312246936985\\
2.44198961530708	-0.712166393311931\\
1.56909217806371	-0.9051528969757\\
0.689876560420579	-1.09449466939142\\
-0.192116946643726	-1.27942929806858\\
-1.07333686657038	-1.45921211652901\\
-1.95023483776291	-1.63311920281601\\
-2.81927990158701	-1.80045029447251\\
-3.67697272029469	-1.96053160825088\\
-4.51985966762235	-2.11271855320024\\
-5.3445467353248	-2.25639832620671\\
-6.14771319964857	-2.39099237953528\\
-6.92612499271441	-2.51595875043732\\
-7.67664772496726	-2.63079424344332\\
-8.39625930625665	-2.7350364565534\\
-9.08206211472713	-2.82826564316701\\
-9.73129466451885	-2.91010640225433\\
-10.3413427252964	-2.98022918996369\\
-10.9097498488316	-3.03835164657839\\
-11.4342272602538	-3.0842397334796\\
-11.9126630741382	-3.11770867553728\\
-12.3431307983233	-3.13862370513442\\
-12.7238970912146	-3.14690060482867\\
-13.0534287413398	-3.14250604646623\\
-13.3303988410499	-3.1254577253826\\
-13.5536921295084	-3.09582428914968\\
-13.7224094834525	-3.05372506115624\\
-13.8358715376457	-2.99932956013477\\
-13.8936214204414	-2.93285681756931\\
-13.8954265934441	-2.85457449573303\\
-13.8412797878592	-2.76479780990672\\
-13.7313990337621	-2.66388825911808\\
-13.5662267821677	-2.55225217051273\\
-13.3464281234368	-2.43033906321822\\
-13.0728881091914	-2.29863983828916\\
-12.7467081885247	-2.15768480202192\\
-12.369201772854	-2.00804153059834\\
-11.9418889472768	-1.85031258465679\\
-11.4664903497241	-1.68513308299313\\
-10.9449202425595	-1.51316814516147\\
-10.3792788045194	-1.33511021327251\\
-9.77184367403535	-1.15167626377354\\
-9.12506077798727	-0.963604920437306\\
-8.44153448281958	-0.771653480184829\\
-7.72401710767657	-0.576594863717932\\
-6.97539784178465	-0.37921450324041\\
-6.19869111070708	-0.180307179799734\\
-5.39702443831609	0.0193261770156987\\
-4.57362585335813	0.218881714137346\\
-3.73181089132146	0.417555891849097\\
-2.87496924394483	0.614548719358121\\
-2.00655111012532	0.809066976075432\\
-1.13005330318504	1.00032740563524\\
-0.249005170438112	1.18755986979187\\
};

\addplot[only marks, mark=*, mark options={}, mark size=2.5pt, color=black!50!green, fill=black!50!green] table[row sep=crcr]{%
x	y\\
0	0\\
};
\addlegendentry{Secondary Sat.}

\addplot[only marks, mark=*, mark options={}, mark size=2.5pt, color=mycolor_1, fill=mycolor_1] table[row sep=crcr]{%
x	y\\
1.9742749372742	-0.0695989240354739\\
};
\addlegendentry{dv limit = 0.004 $\frac{m}{s}$}

\addplot[only marks, mark=*, mark options={}, mark size=2.5pt, color=mycolor_2, fill=mycolor_2] table[row sep=crcr]{%
x	y\\
2.41437308679634	-0.0695989240354739\\
};
\addlegendentry{dv limit = 0.006 $\frac{m}{s}$}

\addplot[only marks, mark=*, mark options={}, mark size=2.5pt, color=mycolor_3, fill=mycolor_3] table[row sep=crcr]{%
x	y\\
2.82598715138624	-0.0695989240354739\\
};
\addlegendentry{dv limit = 0.008 $\frac{m}{s}$}

\addplot[only marks, mark=*, mark options={}, mark size=2.5pt, color=mycolor_4, fill=mycolor_4] table[row sep=crcr]{%
x	y\\
3.21038948393758	-0.0695989240354739\\
};
\addlegendentry{dv limit = 0.01 $\frac{m}{s}$}

\addplot[only marks, mark=*, mark options={}, mark size=2.5pt, color=black!40!orange, fill=black!40!orange] table[row sep=crcr]{%
x	y\\
0.998461159101683	-6.60170702826424e-17\\
};
\addlegendentry{No Maneuver}

\end{axis}

\end{tikzpicture}%
}
    \caption{The maneuver end state in the b-plane for scenarios with different delta-v limits.}
    \label{sdp-manuever-effect-example-3}
  \end{subfigure}
  \hfill
  \begin{subfigure}[t]{0.48\textwidth}
    \centering
    \scalebox{1.0}{
%
%
\definecolor{mycolor1}{rgb}{0.00000,0.44700,0.74100}%
\definecolor{mycolor2}{rgb}{0.85000,0.32500,0.09800}%
\definecolor{mycolor3}{rgb}{0.92900,0.69400,0.12500}%
\definecolor{mycolor4}{rgb}{0.65, 0.3, 0.75}%
\begin{tikzpicture}

\begin{axis}[%
width=0.93\columnwidth,
height=2.0in,
at={(0.758in,0.481in)},
xmin=0,
xmax=1.0,
xlabel style={font=\color{white!15!black}},
xlabel={Orbit Revolutions},
ymin=0,
ymax=0.18,
ylabel style={font=\color{white!15!black}},
ylabel={$\text{Acceleration mm/s}^\text{2}$},
axis background/.style={fill=white},
title style={font=\bfseries},
title={Maneuver Magnitude Contingency Case},
axis x line*=bottom,
axis y line*=left,
scaled y ticks = true,
legend style={at={(0.6,1.03)}, legend cell align=left, align=left, draw=white!15!black, legend columns =1}
]
\addplot [color=mycolor1, line width=2.0pt]
  table[row sep=crcr]{%
0	0.00298051828472869\\
0.0204081632653061	0.00477008257070622\\
0.0408163265306122	0.007165887500619\\
0.0612244897959184	0.00978098146592294\\
0.0816326530612245	0.0125346964359729\\
0.102040816326531	0.0154051510194173\\
0.122448979591837	0.0183818343493413\\
0.142857142857143	0.0214525501065927\\
0.163265306122449	0.0246011246291618\\
0.183673469387755	0.0278061010618076\\
0.204081632653061	0.0310408252948352\\
0.224489795918367	0.0342670597823009\\
0.244897959183673	0.0345789423130934\\
0.26530612244898	0.0345791408689776\\
0.285714285714286	0.0345791607261949\\
0.306122448979592	0.0345791780103784\\
0.326530612244898	0.0345791841378592\\
0.346938775510204	0.0345791915736947\\
0.36734693877551	0.0345791874244211\\
0.387755102040816	0.0345791867707189\\
0.408163265306122	0.0345791909830838\\
0.428571428571429	0.0345791972812362\\
0.448979591836735	0.0345791955428577\\
0.469387755102041	0.0345791932525573\\
0.489795918367347	0.0345791927396583\\
0.510204081632653	0.0345791927968657\\
0.530612244897959	0.0345791935343955\\
0.551020408163265	0.0345791945467086\\
0.571428571428571	0.0345791943135679\\
0.591836734693878	0.0345791933834982\\
0.612244897959184	0.0345791921294014\\
0.63265306122449	0.0345791902510445\\
0.653061224489796	0.03457918685024\\
0.673469387755102	0.0345791798419451\\
0.693877551020408	0.0345791665188909\\
0.714285714285714	0.0345791322272919\\
0.73469387755102	0.0345788388865506\\
0.755102040816327	0.0324869102369063\\
0.775510204081633	0.0292547694789802\\
0.795918367346939	0.0260449410283888\\
0.816326530612245	0.0228900409125324\\
0.836734693877551	0.0198160439348218\\
0.857142857142857	0.0168435966235922\\
0.877551020408163	0.0139839447349321\\
0.897959183673469	0.011240237776628\\
0.918367346938776	0.00860739173032833\\
0.938775510204082	0.00607164303907334\\
0.959183673469388	0.00361112984250602\\
0.979591836734694	0.00119832457644823\\
};
\addlegendentry{dv limit = 0.004 $\frac{m}{s}$}

\addplot [color=mycolor2, line width=2.0pt]
  table[row sep=crcr]{%
0	0.00349595228060391\\
0.0204081632653061	0.00580730416797156\\
0.0408163265306122	0.00883043947148662\\
0.0612244897959184	0.0121082400460858\\
0.0816326530612245	0.0155517592515552\\
0.102040816326531	0.0191382496925539\\
0.122448979591837	0.0228562799345066\\
0.142857142857143	0.0266922561883771\\
0.163265306122449	0.0306266431598579\\
0.183673469387755	0.034632935562383\\
0.204081632653061	0.0386780359351481\\
0.224489795918367	0.0427231635775575\\
0.244897959183673	0.046724976373344\\
0.26530612244898	0.0506372103129528\\
0.285714285714286	0.051868639638446\\
0.306122448979592	0.0518687783962788\\
0.326530612244898	0.0518688122038611\\
0.346938775510204	0.0518688184316977\\
0.36734693877551	0.0518688149584574\\
0.387755102040816	0.051868816891771\\
0.408163265306122	0.0518688125136957\\
0.428571428571429	0.0518688126336528\\
0.448979591836735	0.051868817213996\\
0.469387755102041	0.0518688184778481\\
0.489795918367347	0.0518688184013522\\
0.510204081632653	0.051868818576898\\
0.530612244897959	0.0518688188905401\\
0.551020408163265	0.0518688187713065\\
0.571428571428571	0.0518688184520661\\
0.591836734693878	0.0518688178731704\\
0.612244897959184	0.0518688165165656\\
0.63265306122449	0.0518688146301729\\
0.653061224489796	0.0518688105357005\\
0.673469387755102	0.0518687972227697\\
0.693877551020408	0.0518685679622651\\
0.714285714285714	0.0486650566032044\\
0.73469387755102	0.044699806814429\\
0.755102040816327	0.0406714484153272\\
0.775510204081633	0.0366278210372127\\
0.795918367346939	0.032612092855835\\
0.816326530612245	0.0286639347842044\\
0.836734693877551	0.0248166960342155\\
0.857142857142857	0.0210958705965132\\
0.877551020408163	0.0175153316445848\\
0.897959183673469	0.0140794897397871\\
0.918367346938776	0.0107819417908528\\
0.938775510204082	0.00760584115434226\\
0.959183673469388	0.0045236375497782\\
0.979591836734694	0.00150113302969291\\
};
\addlegendentry{dv limit = 0.006 $\frac{m}{s}$}

\addplot [color=mycolor3, line width=2.0pt]
  table[row sep=crcr]{%
0	0.0039747005466937\\
0.0204081632653061	0.0067767562464497\\
0.0408163265306122	0.0103926869294158\\
0.0612244897959184	0.0142954069703021\\
0.0816326530612245	0.0183886996526084\\
0.102040816326531	0.0226495201965945\\
0.122448979591837	0.0270661682710505\\
0.142857142857143	0.0316235508829652\\
0.163265306122449	0.0362988684324739\\
0.183673469387755	0.0410608353523871\\
0.204081632653061	0.045870228044836\\
0.224489795918367	0.0506809134454414\\
0.244897959183673	0.0554413277014988\\
0.26530612244898	0.0600958766449396\\
0.285714285714286	0.0645864370870164\\
0.306122448979592	0.0688526438471836\\
0.326530612244898	0.069158276809597\\
0.346938775510204	0.0691583539953361\\
0.36734693877551	0.0691583765001639\\
0.387755102040816	0.0691584015769397\\
0.408163265306122	0.0691583908386955\\
0.428571428571429	0.0691583923679152\\
0.448979591836735	0.0691583908128461\\
0.469387755102041	0.0691583918442672\\
0.489795918367347	0.0691583983604276\\
0.510204081632653	0.0691584015883445\\
0.530612244897959	0.0691584015120244\\
0.551020408163265	0.069158399047653\\
0.571428571428571	0.0691583958371186\\
0.591836734693878	0.0691583929539911\\
0.612244897959184	0.0691583857139981\\
0.63265306122449	0.0691583466016742\\
0.653061224489796	0.0691578340089502\\
0.673469387755102	0.0668477086447224\\
0.693877551020408	0.0624659660499211\\
0.714285714285714	0.0578885776156821\\
0.73469387755102	0.0531738205789504\\
0.755102040816327	0.0483847210269442\\
0.775510204081633	0.0435779412812737\\
0.795918367346939	0.0388016249096441\\
0.816326530612245	0.0341050000030325\\
0.836734693877551	0.0295284091092749\\
0.857142857142857	0.0251025181597951\\
0.877551020408163	0.0208433500385021\\
0.897959183673469	0.0167553025568947\\
0.918367346938776	0.012831407331047\\
0.938775510204082	0.00905178981349475\\
0.959183673469388	0.00538348752852062\\
0.979591836734694	0.00178636525897982\\
};
\addlegendentry{dv limit = 0.008 $\frac{m}{s}$}

\addplot [color=mycolor4, line width=2.0pt]
  table[row sep=crcr]{%
0	0.00440245771509798\\
0.0204081632653061	0.0076796837819547\\
0.0408163265306122	0.011854259394161\\
0.0612244897959184	0.0163439795459802\\
0.0816326530612245	0.02104760006958\\
0.102040816326531	0.0259419038118659\\
0.122448979591837	0.0310146990264902\\
0.142857142857143	0.0362496907649417\\
0.163265306122449	0.0416209260443959\\
0.183673469387755	0.0470927787383755\\
0.204081632653061	0.052620207800149\\
0.224489795918367	0.058150222462159\\
0.244897959183673	0.0636235864468053\\
0.26530612244898	0.068976465186614\\
0.285714285714286	0.0741423808503323\\
0.306122448979592	0.0790521668834956\\
0.326530612244898	0.0836402757447581\\
0.346938775510204	0.0864447641342672\\
0.36734693877551	0.0864474904213161\\
0.387755102040816	0.086447899963936\\
0.408163265306122	0.0864479629432624\\
0.428571428571429	0.0864479775132142\\
0.448979591836735	0.0864479842104555\\
0.469387755102041	0.0864479898612456\\
0.489795918367347	0.0864479891559681\\
0.510204081632653	0.0864479851638078\\
0.530612244897959	0.0864479796337913\\
0.551020408163265	0.086447970145753\\
0.571428571428571	0.0864479488627319\\
0.591836734693878	0.0864478649346095\\
0.612244897959184	0.086447498103785\\
0.63265306122449	0.0859805608196807\\
0.653061224489796	0.081604317478912\\
0.673469387755102	0.0768596356713336\\
0.693877551020408	0.0718215691442395\\
0.714285714285714	0.0665588387417584\\
0.73469387755102	0.0611398207706224\\
0.755102040816327	0.0556348323400756\\
0.775510204081633	0.0501081744306319\\
0.795918367346939	0.0446195960715399\\
0.816326530612245	0.0392221257872498\\
0.836734693877551	0.0339593566481107\\
0.857142857142857	0.0288710561889067\\
0.877551020408163	0.0239732704830692\\
0.897959183673469	0.0192718471346196\\
0.918367346938776	0.0147597044821793\\
0.938775510204082	0.0104118244585026\\
0.959183673469388	0.00619247268710601\\
0.979591836734694	0.00205477387840773\\
};
\addlegendentry{dv limit = 0.01 $\frac{m}{s}$}

\end{axis}

\end{tikzpicture}%
}
    \caption{Magnitude of the collision-avoidance maneuvers for scenarios with different delta-v limits.}
    \label{sdp-maneuver-mag-example-3}
  \end{subfigure}
  \caption{(Example 3) The boundary of the ellipse represents all relative positions corresponding to a collision probability of $P_c =1 \times 10^{-6}$. The blue, red, yellow, and purple markers in (a) represent scenarios with different delta-v upper bound limits in which the satellite cannot achieve the desired probability of collision. The computed maneuvers in (b) minimize collision risk and satisfy the specified upper bound limit.}
  \label{sdp-example3-results}
\end{figure*}

\subsection{Example 3}
In this example, we reduce the per-timestep delta-v limit to create scenarios where the satellite lacks sufficient thrust to achieve the desired probability of collision within the available time. In these cases, solving problem \eqref{eq:sdp} leads to a solution that is not tight. Our contingency plan results in a tight relaxation as shown in Figure \ref{tightness-combined}.  
We evaluate four delta-v limits (0.004 $m/s$, 0.006 $m/s$, 0.008 $m/s$, 0.01 $m/s$), and present the results for the change in the b-plane and magnitude of the maneuvers in Figure \ref{sdp-example3-results}. As expected, increasing the delta-v limit enables the spacecraft to achieve a probability of collision closer to the desired value. The solution trades off control effort and collision risk according to the relative weights assigned to each term in the objective function. For this example, $\alpha$ was set to 10.  

In summary, our method reliably computes the globally optimal solution without relying on an initial guess, guarantees convergence, and consistently achieves superior performance compared to alternative approaches.

\subsection{Limitations}
Our approach guarantees a globally optimal solution; however, this advantage comes at the cost of increased problem size. Formulating the problem as an SDP requires introducing additional variables (the moment matrices). Consequently, the number of decision variables grows rapidly with the problem horizon, leading to longer solution times. While the SDP formulation is larger than the original problem, this trade off is outweighed by its reliability and optimality guarantees.

Our formulation also relies on simplifying modeling assumptions. In particular, we assume linearized orbital dynamics, Gaussian uncertainty for both spacecraft, and that the time of closest approach occurs at the end of the planning horizon. These assumptions facilitate the convex problem structure and are justified in practice since the method is intended for short-duration conjunction scenarios. However, if higher model fidelity is desired, a polishing step could be performed using IPOPT with the original nonlinear dynamics, treating time as a decision variable, and using the SDP solution as a warm start. This step is not part of the proposed method but remains a viable option for refinement if needed. 

\section{CONCLUSIONS}
\label{conclusions}
In this work, we introduced a convex semidefinite programming formulation for the collision‑avoidance problem, representing an important advance toward fully autonomous maneuver planning for collision avoidance. By applying Shor's relaxation and empirically demonstrating tightness, we obtain globally optimal solutions without the need for an initial guess --- a capability not offered by existing methods. Our results demonstrate minimum-energy maneuvers that satisfy a probability-of-collision constraint. When the desired probability of collision cannot be met due to thrust or time limitations, our framework produces maneuvers that minimize collision risk. 

Looking ahead, we aim to extend this framework to incorporate the true nonlinear two‑body dynamics directly into the QCQP formulation. We also plan to apply it to other mission‑design problems where initial‑guess dependence remains a major challenge, such as low‑thrust orbit transfers and interplanetary mission design.

\addtolength{\textheight}{-12cm}   






\section*{ACKNOWLEDGMENT}
This material is based upon work supported by the National Science Foundation under Grant No. DGE2140739.


\bibliographystyle{IEEEtran}
\bibliography{bibliography/root}
\end{document}